\documentclass[letterpaper]{article} 
\usepackage{aaai23}  
\usepackage{times}  
\usepackage{helvet}  
\usepackage{courier}  
\usepackage[hyphens]{url}  
\usepackage{amsmath,amssymb}
\usepackage{newtxmath}
\usepackage{diagbox}
\usepackage{booktabs}
\usepackage{multirow}
\usepackage{makecell}
\usepackage{enumitem}
\usepackage{graphicx} 
\usepackage{natbib}  
\usepackage{caption} 
\usepackage{algorithm}
\usepackage{algorithmic}
\newcommand{\eg}{{\emph{e.g.}}}
\newcommand{\etal}{{\emph{et al.}}}
\newcommand{\ie}{{\emph{i.e.}}}
\newcommand{\vect}[1]{\mbox{\boldmath{$#1$}}}

\usepackage{newfloat}
\usepackage{listings}
\DeclareCaptionStyle{ruled}{labelfont=normalfont,labelsep=colon,strut=off} 
\floatstyle{ruled}
\newfloat{listing}{tb}{lst}{}
\floatname{listing}{Listing}
\title{Find Beauty in the Rare: Contrastive Composition Feature Clustering for Nontrivial Cropping Box Regression}

\author {
    Zhiyu Pan\equalcontrib\textsuperscript{\rm 1 },
    Yinpeng Chen\equalcontrib \textsuperscript{\rm 1},
    Jiale Zhang \textsuperscript{\rm 1},
    Hao Lu \textsuperscript{\rm 1},
    Zhiguo Cao \textsuperscript{\rm 1}\thanks{Corresponding author.},
    Weicai Zhong \textsuperscript{\rm 2}
}
\affiliations {
    \textsuperscript{\rm 1} Key Laboratory of Image Processing and Intelligent Control, Ministry of Education School of Artificial Intelligence and Automation, Huazhong University of Science and Technology\\
    \textsuperscript{\rm 2} Huawei CBG Consumer Cloud Service Search Product \& Big Data Platform Department\\
    \{zhiyupan, zgcao\}@hust.edu.cn
}

\usepackage{bibentry}

\begin{document}
\maketitle

\begin{abstract}
Automatic image cropping algorithms aim to recompose images like human-being photographers by generating the cropping boxes with improved composition quality. Cropping box regression approaches learn the beauty of composition from annotated cropping boxes. However, the bias of annotations leads to quasi-trivial recomposing results, which has an obvious tendency to the average location of training samples. The crux of this predicament is that the task is naively treated as a box regression problem, where rare samples might be dominated by normal samples, and the composition patterns of rare samples are not well exploited. Observing that similar composition patterns tend to be shared by the cropping boundaries annotated nearly, we argue to find the beauty of composition from the rare samples by clustering the samples with similar cropping boundary annotations, \ie, similar composition patterns. We propose a novel Contrastive Composition Clustering (\textbf{C2C}) to regularize the composition features by contrasting dynamically established similar and dissimilar pairs. In this way, common composition patterns of multiple images can be better summarized, which especially benefits the rare samples and endows our model with better generalizability to render nontrivial results. Extensive experimental results show the superiority of our model compared with prior arts. We also illustrate the philosophy of our design with an interesting analytical visualization.
\end{abstract}

\begin{figure}[!t]
\centering
\includegraphics[width=0.85\columnwidth]{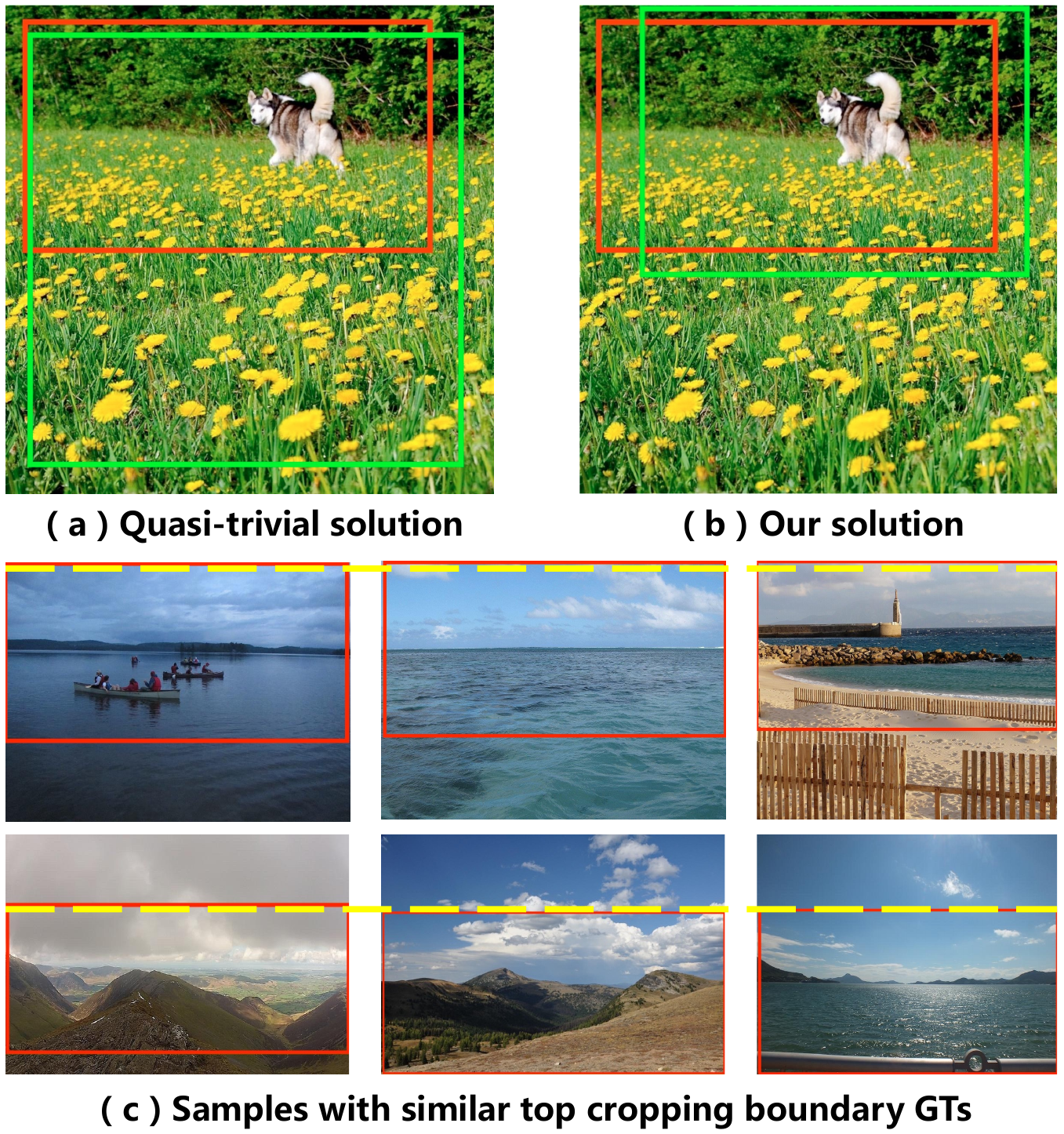}
\vspace{-8pt}
\caption{An illustration of quasi-trivial solution and the examples of our observation. (a) The cropping result from CACNet~\cite{hong2021composing}. (b) The cropping result from our approach. (c) Samples whose top cropping boundaries locate relatively nearly at the original image coordinate. Red and green boxes indicate the annotated and predicted cropping results respectively; Yellow dotted line indicates an identical relative reference position.}
\vspace{-13pt}
\label{fig:fig1}
\end{figure}

\section{Introduction}
Automatic image cropping is an effective technology to improve composition quality of the input images. As one of the key components of computational imaging, it has attracted wide research interest ~\cite{chen2003visual,islam2017survey}. 
Provided human-annotated cropping boxes~\cite{wang2018deep,zeng2019reliable}, automatic cropping can be formalized as a cropping box regression problem. However, as regression models are easily affected by the bias of training set~\cite{yang2021delving}, recent regression-based approaches~\cite{hong2021composing,pan2021robust} are dominated by the normal samples and tend to render quasi-trivial cropping boxes (Fig.~\ref{fig:fig1}(a)). The quasi-trivial cropping boxes can be described as the central cropping boxes whose boundaries locate near the average location of training samples. When the quasi-trivial cropping boxes just inherit the location bias of the annotated cropping boundaries and ignore the composition patterns of the input images, the improvement of composition quality is little, which limits the real-world application. Trying to generate nontrivial cropping boxes, previous methods take extra data with additional annotations, \eg, composition classification dataset~\cite{lee2018photographic}, into account. However, the predicament is hardly addressed. Therefore, we try to reveal the intrinsic cause of quasi-trivial solution. As shown in Fig.~\ref{fig:motivation_analysis}, by analysing the CACNet baseline~\cite{hong2021composing}, we illustrate the cause of quasi-trivial solution: \emph{dominated by the normal samples, the composition patterns of rare samples are barely exploited}. Hence, we argue to find the beauty of composition by taking full advantage of rare samples rather than by introducing extra data.

In this paper, we probe into the problem: \emph{how to better exploit the composition patterns of the rare training samples?} Similar problems have been studied for decades by the imbalanced learning methods~\cite{branco2017smogn, lin2017focal,yang2021delving}, whose main idea is re-weighting. However, directly adapting these methods into the cropping box regression task can barely improve the composition quality. Therefore, we delve into this problem from another view, which is to enhance the composition representations of the rare by looking for the connections between samples. As illustrated in Fig.~\ref{fig:fig1}(c), compared in the same row, when the annotated top cropping boundaries tend to locate at a close relative location, the visual focuses locate relatively nearly; compared with two different rows, the focuses locate relatively differently. Based on this observation, we make a hypothesis: \emph{cropping boundaries with close relative location annotations share the similar composition patterns and vice versa}. Based on this, we make full use of the relationships between the samples, which especially benefits the rare ones.

Inspired by the contrastive learning~\cite{hadsell2006dimensionality,chen2020simple,he2020momentum} and deep clustering works~\cite{zhou2022comprehensive,10.1007/978-3-031-19818-2_14}, we propose Contrastive Composition Clustering (\textbf{C2C}) to regularize the composition features based on their regression target distance. Due to that our hypothesis is based on the cropping boundary, each image has four composition features to depict each cropping boundary respectively. The optimization targets of C2C are: (1) draw features of the samples with similar annotations closer (2) 
widen the feature distance of negative pairs where annotated boundary locates far away from each other. When C2C is performed in the mini-batch, we calculate the distance map to select positive and negative pairs dynamically. Based on the distance map, after breaking the contrastive loss into alignment and uniformity loss~\cite{wang2020understanding}, the contrastive clustering procedure can successfully work in our regression scenario. As illustrated in Fig.~\ref{fig:fig1}(a) and (b), benefited from the proposed C2C, our solution can render nontrivial cropping boxes.

Extensive experiments on public benchmarks show that our solution outperforms state-of-the-art regression-based approaches even without extra data. By visualizing the cropping results, with C2C, our model hardly collapses into quasi-trivial solutions. Compared with other imbalanced learning baselines, C2C also shows superiority in both performance and time efficiency. For the first time, without external data, we demonstrate state-of-the-art image cropping with only a deep clustering procedure.

\section{Related Work}
\subsection{Image Cropping}
Conventional image cropping algorithms~\cite{chen2003visual,suh2003automatic,marchesotti2009framework,cheng2010learning,greco2013saliency} formulate the composition patterns with predefined rules~\cite{zhang2005auto,nishiyama2009sensation,yan2013learning,fang2014automatic}. However, performance of the classic algorithms are far from satisfactory. Recently, data-driven algorithms have significantly improvement. They are driven by two main ideas: candidate box selection and cropping box regression. 

\textbf{Candidate box selection algorithms} follow a two-stage pipeline. First, candidate boxes are generated according to prior knowledge. Then, the candidate boxes are ranked based on
the learned composition aesthetic knowledge. The knowledge can be modeled by saliency detection~\cite{wang2018deep,tu2020image}, teacher-student architecture~\cite{wei2018good}, region of interest and discard~\cite{zeng2019reliable,zeng2020grid}, view-wise mutual relation~\cite{li2020composing}, visual elements dependencies~\cite{pan2021transview,9745054} and view-wise difference\cite{pan2022discriminate}. But the results 
are dependent on the candidates. To overcome this limitation, some works attempt to find end-to-end solutions.

\textbf{Cropping box regression algorithms} are end-to-end solutions that imitate the process of human cropping. Reinforcement learning algorithms estimate the final cropping box step by step~\cite{li2018a2,li2019fast}. Other algorithms predict the cropping results according to a salient cluster~\cite{pan2021robust}, or a composition class~\cite{hong2021composing}. The common deficiencies of these algorithms are the need for extra data and the problem of degenerating into quasi-trivial solutions. In contrast to these approaches, our work can render nontrivial cropping boxes without extra data.
\vspace{-5pt}
\subsection{Contrastive Learning and Deep Clustering}
Motivated by the InfoMax principle~\cite{linsker1988self}, contrastive learning aims to learn representations by contrasting positive pairs against negative pairs~\cite{hadsell2006dimensionality,dosovitskiy2014discriminative}. The pioneer works introduce the memory bank to store the representation vectors and update them by contrasting~\cite{wu2018unsupervised,zhuang2019local,he2020momentum}. Some other works contrast in batches~\cite{doersch2017multi,ji2019invariant,ye2019unsupervised,chen2020simple}. Recently, Wang~\etal~\cite{wang2020understanding} has characterized the contrastive learning by alignment and uniformity. 
When deep clustering~\cite{zhou2022comprehensive} concerns the joint optimization of clustering and deep representation learning, some works~\cite{ling2022vision,deng2022strongly,chen2022design} introduce the idea of contrasting into deep clustering. 
In this work, we also conduct composition feature clustering based on the idea of contrasting and combine the clustering procedure into regular training process, which helps to get a better composition representation.
\vspace{-5pt}
\subsection{Imbalanced Learning}
In confront of the training data bias, a naive idea is to re-weight the loss with the frequency of samples~\cite{lin2017focal}. Some other works also try to create new samples to balance the original distribution~\cite{torgo2013smote,branco2017smogn} or to decouple the encoder and decoder~\cite{kang2019decoupling}. Recently, Yang~\etal~\cite{yang2021delving} proposes the label and feature distribution smoothing to train the regressor unbiasedly. We directly adapt these methods into our problem as baselines and compare our C2C with them in the experiment section.

\label{sec:motivation}
\begin{figure*}[!t]
\centering
\includegraphics[width=0.9\textwidth]{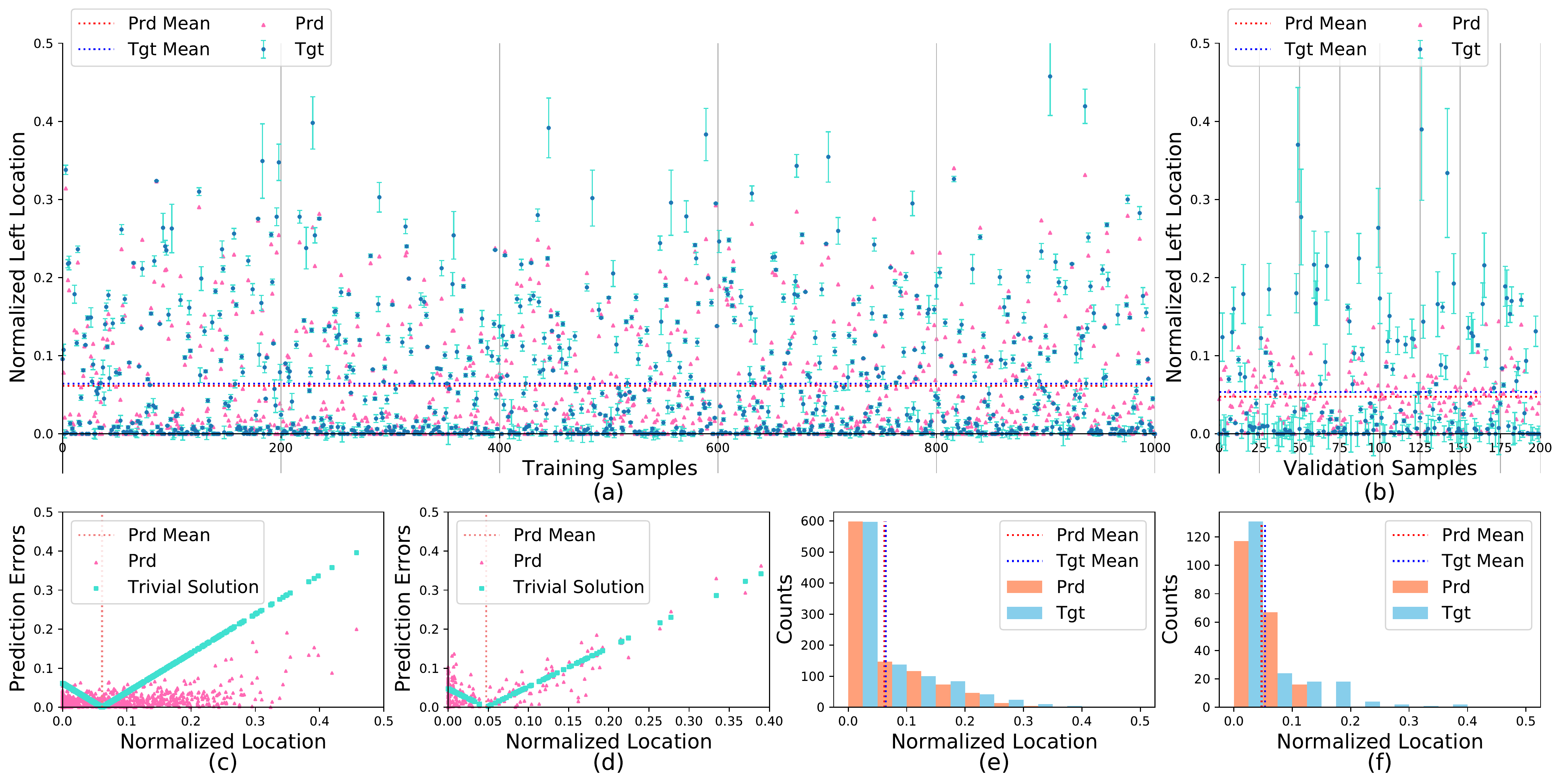}
\vspace{-10pt}
\caption{Statistical analysis of the left cropping boundaries generated from CACNet vanilla regression baseline. (a) The test results on the training set. (b) The test results on the validation set. (c) The error distribution on the training set. (d) The error distribution on the validation set. (e) The location distribution on the training set. (f) The location distribution on the validation set. ``prd" means the prediction, ``tgt" denotes the target. Best viewed in color.}
\vspace{-10pt}
\label{fig:motivation_analysis}
\end{figure*}

\section{Problem and Approach}
\subsection{Quasi-Trivial Solution}

We explain what is the quasi-trivial cropping solution and illustrate the cause of it by analyzing the recomposed results of the vanilla regression baseline CACNet~\cite{hong2021composing}. To make the analysis unaffected by extra data, the composition classification module is dropped. 
Statistic of the left boundary is shown in Fig.~\ref{fig:motivation_analysis} (results of other boundaries are in similar distributions).
The red and blue points in Fig.~\ref{fig:motivation_analysis}(a-b) show the predicted and target left boundary location respectively. The location is normalized into $[0,1]$, where $0$ represents the leftmost location and $1$ represents the rightmost location. The ordinal number of samples are indicated by the horizontal coordinate. The error bar on the target point colored in light blue indicates the absolute value of the prediction error on this sample. From Fig.~\ref{fig:motivation_analysis}(b), on the validation set, we observe that the samples whose target is far away from the mean location tend to have larger prediction error, which means that the baseline tends to locate the boundary at an approximate average location for an unseen image. Therefore, we compare the baseline with the mean location which can serve as a trivial solution. As illustrated in Fig.~\ref{fig:motivation_analysis}(c-d), when the vertical and horizontal coordinates of each point represent the prediction error and the corresponding target location respectively, the trivial solution behaves as two symmetrical lines with the mean location as the axis of symmetry. From Fig.~\ref{fig:motivation_analysis}(d), on the validation set, we find that the baseline acts extremely similar to the trivial solution. We call this the quasi-trivial solution of this boundary. By plotting the location distribution of the targets and predicted results, we find that normal training samples with the most amount locate near the mean location (Fig.~\ref{fig:motivation_analysis}(e)), hence the baseline tends to predict around the mean location for an unseen image (Fig.~\ref{fig:motivation_analysis}(f)), which means the quasi-trivial solution comes from that the normal samples overwhelm the rare samples. Hence, we try to exploit the composition patterns of the rare samples better.
\subsection{Cropping Boundary Locating Network}

\begin{figure*}[!t]
\centering
\includegraphics[width=0.9\textwidth]{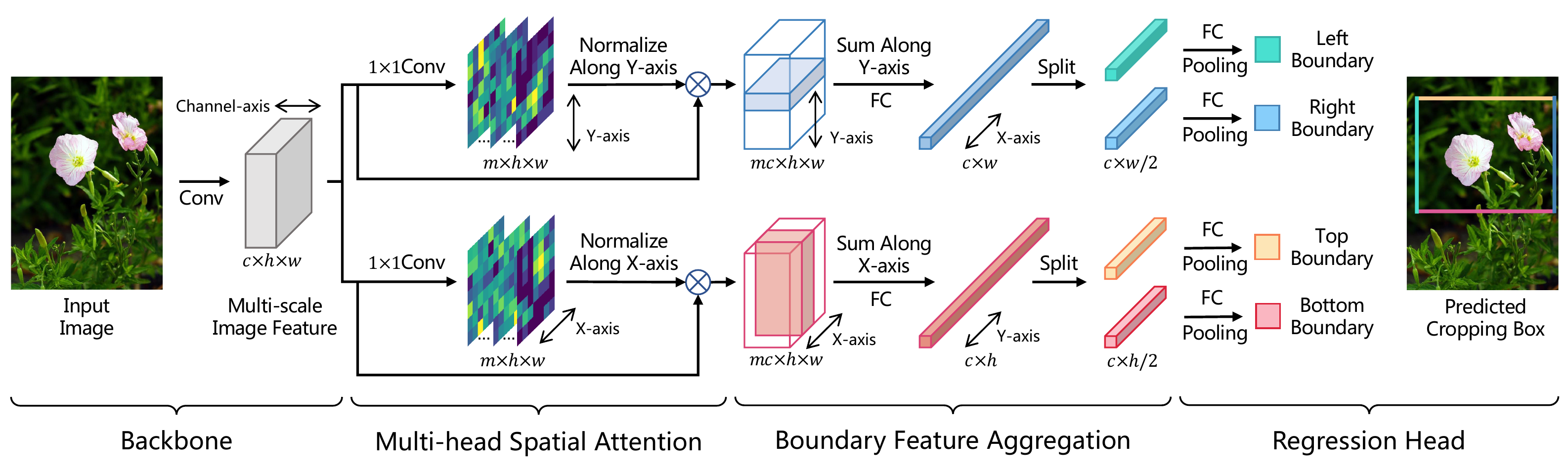}
\vspace{-10pt}
\caption{The pipeline of the cropping boundary locating network. The multi-scale features extracted from the input image are used to encode the horizontal and vertical features. Then, the horizontal and vertical features are split into boundary features. The final cropping boxes are predicted based on the four boundary features.}
\vspace{-10pt}
\label{fig:pipeline}
\end{figure*}
As the imbalanced distributions of four cropping boundaries have different levels and forms, the whole cropping box regression problem is divided into four boundary locating subtasks. Inspired by SABL~\cite{wang2020side}, we build the cropping boundary locating network (CBLNet). As shown in Fig.~\ref{fig:pipeline}, CBLNet contains three main components: a convolutional backbone, a boundary feature encoder which builds the boundary features, and a group of regression heads that predict the location of four cropping boundaries.

\textbf{Backbone.} A MobileNetV2-based~\cite{sandler2018mobilenetv2} backbone that fuses multi-scale features~\cite{zeng2019reliable,zeng2020grid} is employed. It generates multi-scale feature maps $\boldmath{\mathcal{F}}\in\vmathbb{R}^{C\times H\times W}$ from the original image, where $C$ is the channel dimensionality, $H$ and $W$ represents the height and width of the feature maps.

\textbf{Boundary feature encoder.} The extracted features $\boldmath\mathcal{F}$ are processed by two multi-head attention modules to encode the horizontal and vertical information. The multi-head attention module is implemented by $m$ $1\times1$ convolutions, where $m$ is the number of heads. The horizontal feature map $\boldmath{\mathcal{H}} \in \vmathbb{R}^{mC \times H \times W}$ and vertical feature map $\boldmath{\mathcal{V}} \in \vmathbb{R}^{mC \times H \times W}$ can be computed by
\begin{gather}
\label{equ:attention}
\boldmath{\mathcal{H}} = \sigma_{y}(\xi(\tau_{0}^{h}(\boldmath{\mathcal{F}})*\boldmath{\mathcal{F}}, \ldots, \tau_{m-1}^{h}(\boldmath{\mathcal{F}})*\boldmath{\mathcal{F}}))\,,\\
\boldmath{\mathcal{V}} = \sigma_{x}(\xi(\tau_{0}^{v}(\boldmath{\mathcal{F}})*\boldmath{\mathcal{F}}, \ldots, \tau_{m-1}^{v}(\boldmath{\mathcal{F}})*\boldmath{\mathcal{F}}))\,,
\end{gather}
where 
$\sigma_y(\cdot)$ and $\sigma_x(\cdot)$ are the normalizing operators along Y and X axis, respectively, $\xi$ is the concatenating operator along the channel axis, 
$\tau_{i}^{h}(\cdot)$ and $\tau_{i}^{v}(\cdot)$ represent the $i$-th head $1\times1$ convolution of the horizontal and vertical attention modules, respectively.
After the dimension reduction and the sum operations along the Y and X axis, $\boldmath{\mathcal{H}}$ and $\boldmath{\mathcal{V}}$ are transformed into the horizontal vector $\vect h \in \vmathbb{R}^{C \times W}$ and vertical vector $\vect v \in \vmathbb{R}^{C \times H}$, respectively. Then, $\vect h$ and $\vect v$ are split into boundary features $\vect x^l,\vect x^r \in \vmathbb{R}^{C\times W/2}$ and $\vect x^t,\vect x^b \in \vmathbb{R}^{C\times W/2}$, which are composition features for left, right, top, and bottom cropping boundaries, respectively.

\textbf{Regression head.}
The normalized locations w.r.t.\ the input image of four cropping boundaries are predicted by a pooling operator and a $3$-layer fully-connected network with the $\tt sigmoid$ activation function. All regression heads are supervised with the $\ell_1$ loss.

\subsection{Contrastive Composition Clustering}

\begin{figure}[!t]
\centering
\includegraphics[width=0.9\columnwidth]{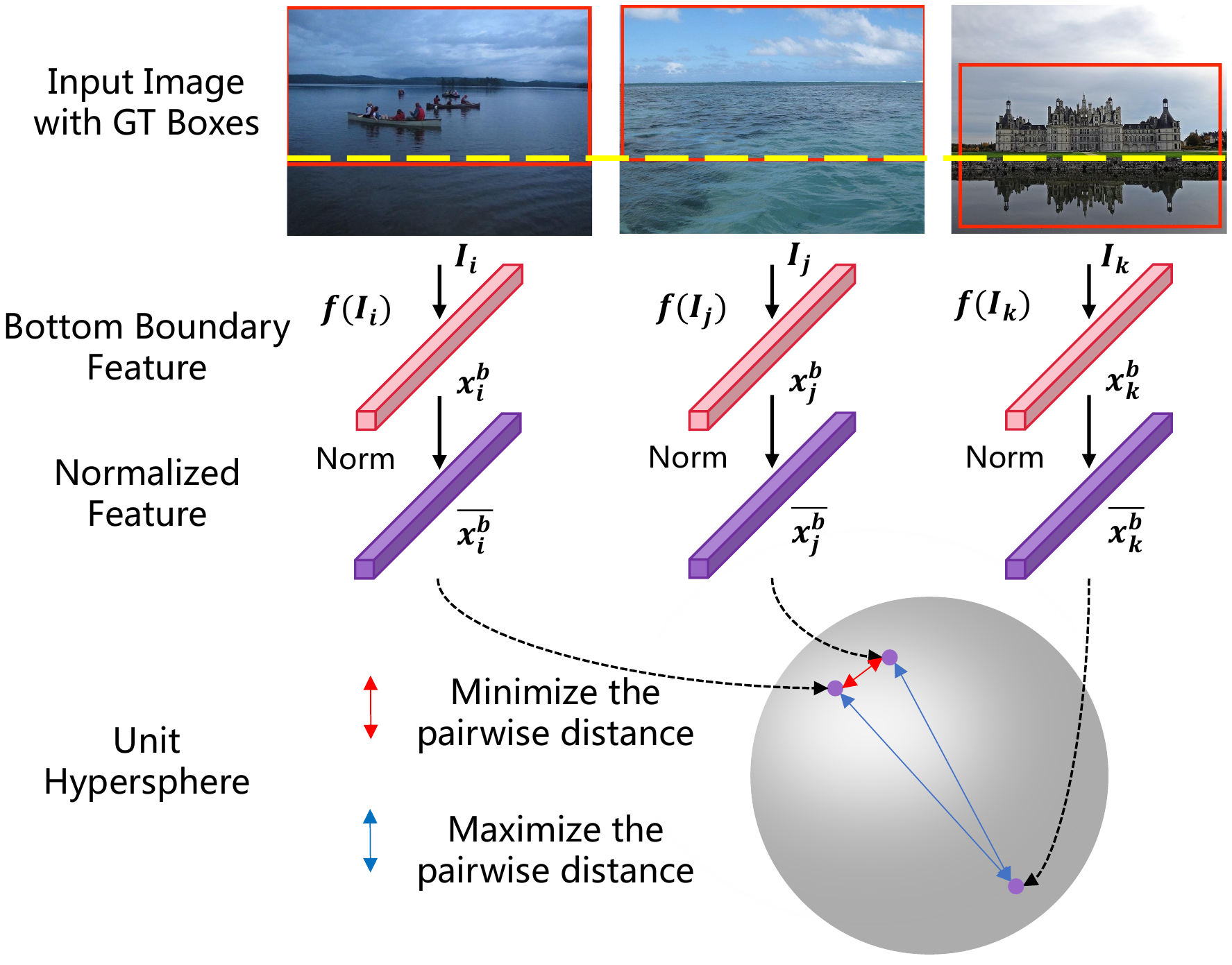}
\vspace{-5pt}
\caption{Contrastive composition clustering. C2C regularizes the features of the cropping boundaries. Taking the learning process of bottom boundary features as an example: the boundary features are $\ell_2$-normalized to the unit hypersphere. The pairwise distances on the unit hypersphere are maximized or minimized conditioned on the pairwise distances of their ground truth boundaries. After training, $\ell_2$ normalization is removed.}
\vspace{-10pt}
\label{fig:contrast}
\end{figure}

\begin{figure*}[!t]
\centering
\includegraphics[width=0.9\textwidth]{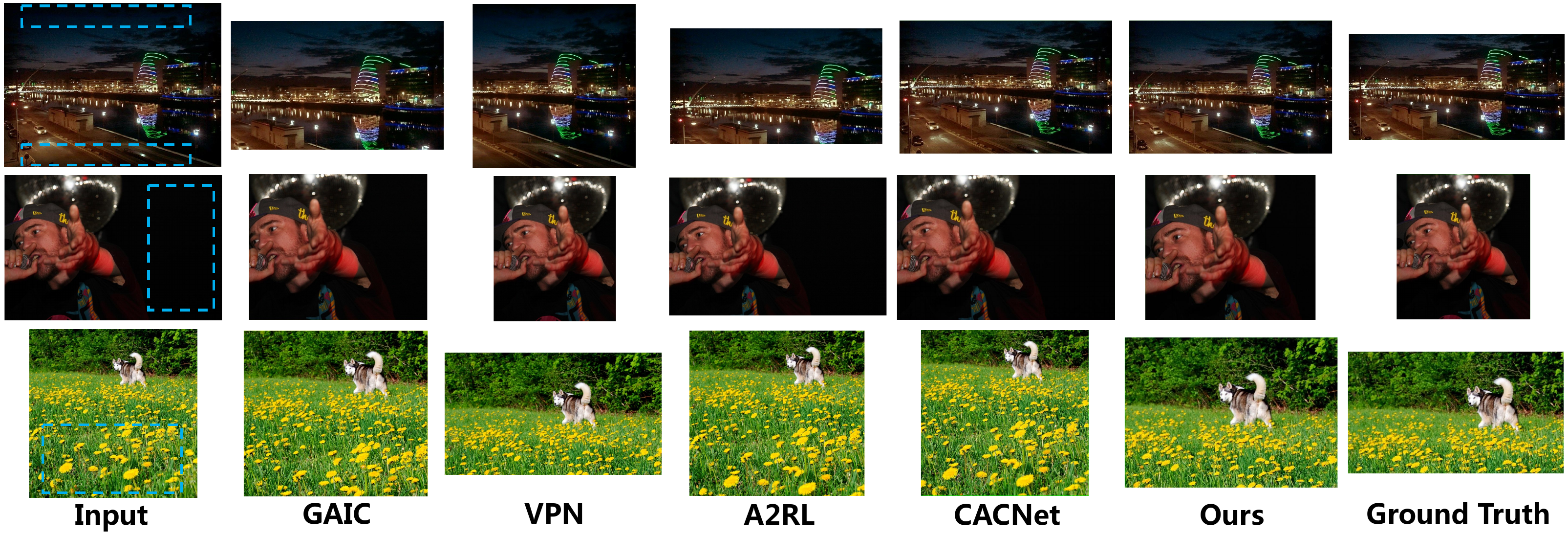}
\vspace{-8pt}
\caption{Qualitative comparison with image cropping models. Redundancy in blue boxes violates the composition.}
\vspace{-10pt}
\label{fig:comparison}
\end{figure*}

After dividing the whole cropping box regression into four sub-regression, composition features w.r.t.\ four cropping boundaries are obtained. Based on our hypothesis, when similar cropping boundaries have common knowledge and dissimilar ones tend to have different composition patterns, it is intuitive to minimize the distance of similar samples and to maximize the distance of dissimilar samples in the composition feature space. As illustrated in Fig.~\ref{fig:contrast}, inspired by the contrastive learning algorithms~\cite{wu2018unsupervised,zhuang2019local,he2020momentum}, we propose the Contrastive Composition Clustering (C2C). 

The C2C introduces the idea of contrastive learning into our regression scenario by setting the positive and negative pair thresholds, \ie, $z_p$ and $z_n$. The positive and negative pair refers to the composition feature pair whose boundary annotations locate nearly and far away from each other, respectively. In this setting, for each batch, the positive and negative pairs are not fixed. Hence, we structure the pairs with a distance map, which is used to calculate the positive and negative pair matrix $\mathcal{P}$ and $\mathcal{N}$. Inspired by \cite{wang2020understanding}, which identifies two contrastive loss properties: alignment and uniformity, we adapt the alignment and uniformity loss into our C2C. The alignment loss is to cluster positive pairs, and the uniformity loss is to discriminate between the elements in negative pairs. In our approach, the regression alignment loss and uniformity loss take the form
\begin{gather}
\label{equ:contrastiveloss}
L_{align} = \sum_{d=l,r,t,b}\underset{0\le i < j\le N}{\vmathbb{E}}\mathcal{P}_{i,j}^d\max(||\overline{\vect x_i^d}-\overline{\vect x^d_j}||_{2}^{\alpha}-\epsilon,0)\\
L_{uniform} = \sum_{d=l,r,t,b}\underset{0\le i < j\le N}{\vmathbb{E}}\mathcal{N}_{i,j}^{d}e^{-t||\overline{\vect x_i^d}-\overline{\vect x^d_j}||_2^2}\,,
\end{gather}
where $\overline{\vect x}$ is the $\ell_2$ normalized boundary composition features, $l,r,t,b$ indicates left, right, top, and bottom respectively, $N$ denotes the batch size, $\epsilon$ controls the expected similarity level of the features with similar regression targets, $\mathcal{P}_{i,j}^d$ and $\mathcal{N}_{i,j}^d$ can be calculated as
\begin{gather}
\mathcal{P}_{i,j}^d = \vmathbb{1}(y_i^d-y_j^d < z_p)\\
\mathcal{N}_{i,j}^d = \vmathbb{1}(y_i^d-y_j^d > z_n)\,,
\end{gather}
where $y$ is the normalized ground truth location of the cropping boundary, and $\vmathbb{1}(\cdot)$ is the sign function.

Details about the derivation process of the optimization target and the pseudo-code of C2C is described in the supplementary. The alignment and uniformity loss are jointly optimized with the $\ell_1$ loss for the cropping boundary locating. The total loss of the cropping boundary regression pipeline can be computed by
\begin{equation}
\label{equ:totalloss}
L = L_1 + \beta L_{align} + \gamma L_{uniform}\,,
\end{equation}
where $\beta$ and $\gamma$ are the hyper-parameters used to control the relative importance of the alignment and uniformity loss. With C2C, we get more informative and discriminative composition features from the rare samples which prevents the model collapse into the quasi-trivial solution.

\section{Experiments}
Here we demonstrate that, CBLNet with C2C achieves better performance than other cropping box regression algorithms with extra data. When imbalanced learning approaches can serve as naive baselines to exploit the rare samples, we compare C2C with these baselines. The results illustrate the superiority of C2C. A detailed ablation study is also conducted to illustrate the effect of each component. Further analyses and visualizations also prove the rationale of our hypothesis and design. 

\subsection{Implementation Details}
The input images are resized to $224\times224$. 
After processed by the backbone pre-trained on ImageNet, the channel dimensionality of multi-scale feature map $\boldmath{\mathcal{F}}$ of size $H=W=14$ is reduced to $C=256$. The head number of the multi-head spatial attention module is set to $m=6$. In the training stage, $32$ images are batched as the input. In the alignment loss, $z_l$ is set to $0.05$, $\alpha$ is set to $1$, and $\epsilon$ is set to $0.5$. In the uniformity loss, $z_h$ is set to $0.7$ and $t$ is set to $1$. Only random cropping is used in data augmentation. By setting $\beta=\gamma=0.025$, the network is optimized by Adam with the learning rate of $5\times10^{-4}$ for $80$ epochs.

\begin{table*}[!t]
\begin{center}
\begin{tabular*}{0.9\textwidth}{@{}@{\extracolsep{\fill}}lcccccccc@{}}
\toprule
\multirow{2}*{\diagbox{Algorithm}{Metric}} &\multicolumn{4}{c}{IoU$\uparrow$} &\multicolumn{4}{c}{BDE$\downarrow$}\\
 &All &Many &Med. &Few &All &Many &Med. &Few\\
\hline
\multicolumn{9}{l}{candidate box selection algorithms}\\
\hline
VPN$^\dag$ &0.665 &\textbf{0.735} &0.631 &\textbf{0.435} &0.085 &\textbf{0.068} &0.092 &\textbf{0.147}\\
GAIC$^\dag$ &0.666 &0.721 &\textbf{0.655} &0.408 &0.084 &0.072 &\textbf{0.086} &\textbf{0.147}\\
TransView &0.682 &- &- &- &0.080 &- &- &-\\
VEN &\textbf{0.735} &- &- &- &\textbf{0.072} &- &- &-\\
\hline
\multicolumn{9}{l}{cropping box regression algorithms}\\
\hline
A2RL$^\dag$ &0.636 &0.734 &0.577 &0.349 &0.097 &0.070 &0.111 &0.186\\
A3RL &0.696 &- &- &- &0.077 &- &- &-\\
vanilla-CACNet$^\ddag$ &0.700 &0.799 &0.650 &0.369 &0.075 &0.050 &0.086 &0.167\\
CACNet$^\ddag$ &0.716 &\textbf{0.809} &0.673 &0.386 &0.070 &\textbf{0.047} &0.079 &0.159\\
regression baseline &0.692 &0.776 &0.655 &0.389 &0.077 &0.056 &0.085 &0.158\\
CBLNet - Ours &0.700 &0.787 &0.659 &0.391 &0.075 &0.052 &0.083 &0.158\\
CBLNet+C2C - Ours &\textbf{0.718} &0.805 &\textbf{0.680} &\textbf{0.418} &\textbf{0.069} &\textbf{0.047} &\textbf{0.078} &\textbf{0.146}\\
\bottomrule
\end{tabular*}
\caption{Quantitative comparison with other state-of-the-art image cropping models on the FCDB dataset. Best performance is in boldface. The performances tagged by $\dag$ are our reproduced results, and $\ddag$ denotes the results from the original authors.}
\label{table:comp_w_crop}
\end{center}
\end{table*}

\begin{table*}[!t]
\begin{center}
\begin{tabular*}{0.9\textwidth}{@{}@{\extracolsep{\fill}}lccccccccc@{}}
\toprule
\multirow{2}*{\diagbox{Algorithm}{Metric}} &\multicolumn{4}{c}{IoU$\uparrow$} &\multicolumn{4}{c}{BDE$\downarrow$} &\multirow{2}*{\makecell[c]{Training\\Time$\downarrow$}}\\
 &All &Many &Med. &Few &All &Many &Med. &Few\\
\hline
CBLNet baseline &0.700 &0.787 &0.659 &0.391 &0.075 &0.052 &0.083 &0.158 &0.720\\
\hline
SMOGN &0.702 &0.796 &0.658 &0.374 &0.075 &0.051 &0.085 &0.166 &0.956\\
RRT &0.701 &0.791 &0.653 &0.416 &0.074 &0.051 &0.086 &0.149 &0.730$\times2$\\
FOCAL-R &0.708 &0.797 &0.663 &0.408 &0.072 &0.050 &0.082 &0.151 &0.747\\
INV &0.707 &0.798 &0.660 &0.404 &0.073 &0.050 &0.083 &0.153 &\textbf{0.734}\\
LDS\&FDS &0.709 &0.798 &0.666 &0.408 &0.072 &0.049 &0.082 &0.151 &1.187\\
C2C &\textbf{0.718} &\textbf{0.805} &\textbf{0.680} &\textbf{0.418} &\textbf{0.069} &\textbf{0.047} &\textbf{0.078} &\textbf{0.146} &0.788\\
\bottomrule
\end{tabular*}
\caption{Quantitative comparison of imbalanced learning algorithms on the FCDB dataset. Best performance is in boldface. The metric of training time shows how many seconds an algorithm needs for training a batch of data. Note that the regressor re-training (RRT) algorithm needs an extra training round, hence the training time is doubled.}
\label{table:comp_w_DIR}
\vspace{-10pt}
\end{center}
\end{table*}

\subsection{Datasets and Evaluation Metrics}
FCDB~\cite{chen2017quantitative} dataset is used for evaluation. The FCDB contains $1743$ images, and each of them has a ground truth cropping box. In the performance comparison, $1395$ images of the FCDB are used for training and $348$ images for testing. We follow the setting of \cite{hong2021composing} to randomly choose $200$ images as the validation set and the other images are used for training. 
To demonstrate whether an approach can render nontrivial cropping boxes, according to their frequency, the test set or the validation set is divided into three disjoint subsets: many-shot, medium-shot, and few-shot samples. The frequency of samples is related to their ground truth box size ratios w.r.t.\ the original image. Hence, in our setting, the ground truth box size ratios of the many-shot, medium-shot, and few-shot samples are in the range of $65\% \sim 100\%$, $40\% \sim 65\%$, and $0\% \sim 40\%$, respectively. The performance of intersection over union (IoU) and the boundary displacement error (BDE) are reported.

\subsection{Performance Comparison}

\subsubsection{Comparison with image cropping models.} Quantitative results on the FCDB dataset are illustrated in Table~\ref{table:comp_w_crop}. For the selection-based algorithms, the recalled top-$1$ cropping boxes are used to calculate the IoU and BDE metrics. Based on the results, we can make the following observations:
(a) \textit{The gap between two types of algorithms lies in the few-shot samples.} This might be a manifestation of the quasi-trivial solution problem. The CBLNet trained with the C2C outperforms the previous regression algorithms especially on the few-shot samples without extra data, which bridges the performance gap between regression-based and selection-based algorithms.
(b) \textit{C2C is more helpful than extra data.} Compared to the vanilla CACNet without the composition classification to the final CACNet, it is obvious that the performance boost is from the extra data with the annotation about composition classification. When the performance of CBLNet without C2C is comparable with vanilla CACNet, after trained with C2C, the CBLNet can perform better than the CACNet, which shows the superiority of the proposed C2C. Qualitative comparison is shown in Fig.~\ref{fig:comparison}. We observe that the selection-based algorithms, \ie, the VPN, can generate nontrivial cropping results but can not obey the symmetry (first row) or visual balance (third row) composition rules well. This shows the limitation of selection-based algorithms. While previous regression-based algorithms tend to render quasi-trivial cropping results, our approach can render nontrivial cropping results and align better with composition rules.

\begin{table*}[t]
\begin{center}
\begin{tabular*}{0.9\textwidth}{@{}@{\extracolsep{\fill}}lcccccccc@{}}
\toprule
\multirow{2}*{\diagbox{Batch Size}{Metric}} &\multicolumn{4}{c}{IoU$\uparrow$} &\multicolumn{4}{c}{BDE$\downarrow$}\\
 &All &Many &Med. &Few &All &Many &Med. &Few\\
\hline
4 &0.734 &0.811 &0.685 &0.424 &0.064 &\textbf{0.041} &0.076 &0.161\\
8 &0.738 &0.805 &0.698 &0.436 &0.061 &0.043 &0.071 &0.145\\
16 &0.741 &0.801 &0.710 &0.438 &0.060 &0.043 &0.066 &0.144\\
32 &\textbf{0.753} &0.809 &0.715 &0.460 &\textbf{0.059} &0.042 &\textbf{0.065} &0.139\\
64 &0.751 &0.809 &0.714 &0.463 &\textbf{0.059} &0.042 &0.066 &0.142\\
128 &0.749 &\textbf{0.813} &0.704 &0.465 &0.062 &0.042 &0.073 &0.142\\
256 &0.750 &0.799 &\textbf{0.718} &\textbf{0.471} &0.061 &0.046 &0.068 &\textbf{0.128}\\
\bottomrule
\end{tabular*}
\caption{Ablation study on the batch size. Best performance is in boldface.}
\label{table:ablation}
\vspace{-10pt}
\end{center}
\end{table*}

\subsubsection{Comparison with imbalanced learning baselines.} The imbalanced learning baselines to be compared:
\begin{itemize}[leftmargin=*]
\item \textit{SMOGN}~\cite{branco2017smogn} defines the rare samples and creates synthetic samples by interpolating the feature and the regression target. During the training stage, Gaussian noise is added to the target, hence, annotated boundary may shift in the range of $10\%$ of the original length or width.

\item \textit{FOCAL-R} is the regression version of the focal loss~\cite{lin2017focal}. Following \cite{yang2021delving}, we rewrite the focal loss as $1/n\sum_{i=1}^{n}\rho(|\mu e_i|)^{\psi}e_i$, where $e_i$ is the $\ell_1$ error for $i$-th sample, $\rho(\cdot)$ is the $\tt sigmoid$ function, and $\mu$ and $\psi$ are set to $\mu=\psi=2$.

\item \textit{Inverse reweighting (INV)} is a traditional method re-weighting the loss according to inverse frequency of the classes. \cite{yang2021delving} adopts the scheme based on the regression target distribution in the regression scenario.

\item \textit{Regressor re-training (RRT)}~\cite{kang2019decoupling} decouples the training process of feature and classifier. \cite{yang2021delving} transforms it into the regression version, re-training the regression head with inverse reweighting.

\item \textit{LDS\&FDS}~\cite{yang2021delving} propose to smooth the distribution of regression target to reweight the loss and recalibrate the statistics of the feature distribution.

\end{itemize}

The results are illustrated in Table~\ref{table:comp_w_DIR}. It shows that previous methods have notable limitations: (a) RRT can boost the performance on the few-shot samples but harms that on the medium-shot samples. (b) LDS\&FDS can improve the performance on many-, medium-, few-shot samples, but at a cost of additional time consumption.
C2C obtains significant gains on all groups of samples with low time cost.

\subsection{Ablation Study}

\begin{figure}[!t]
\centering
\includegraphics[width=0.98\columnwidth]{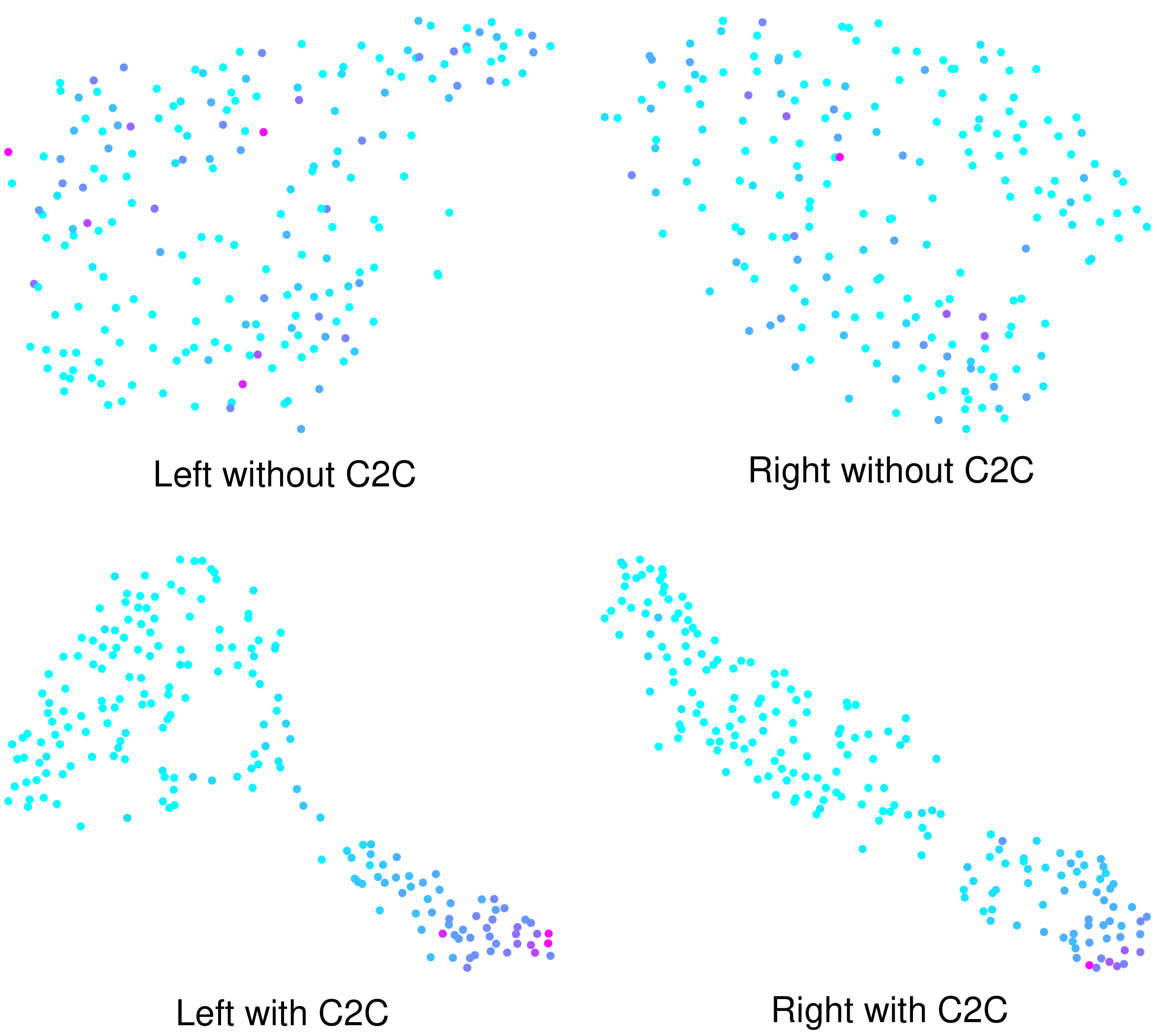}
\vspace{-7pt}
\caption{Distributions of the boundary features from the CBLNet with/without C2C on the validation set. The features are reduced into two dimensions by t-SNE. The color of the points represents the location of their ground truth boundary, the lighter the color is, the closer the ground truth boundary locates to the original image boundary. Best viewed in color.}
\vspace{-10pt}
\label{fig:distribution}
\end{figure}

\subsubsection{Boundary feature encoder and C2C.} The performance contributions of the boundary feature and C2C are shown in Table~\ref{table:comp_w_crop}. We can observe that the regression baseline tends to output the quasi-trivial cropping results. By only adding the boundary feature encoder to build the CBLNet, the performance cannot be significantly improved. After training with C2C, the cropping results align better with the ground truth. This shows that the main contribution of this work is from the C2C rather than the CBLNet design.

\subsubsection{Batch size.} The proposed C2C contrasts the samples in the batch without the memory bank. We vary the batch size from $4$ to $256$. Larger batch size provides more positive and negative pairs. From the results in Table~\ref{table:ablation}, it can be seen that the change of the performance on the many-shot samples is relatively smooth. As for the few-shot samples, with larger batch size, the performance is better. The results reveal that the proposed C2C benefits from large batch sizes, which is consistent with the conclusion in \cite{chen2020simple,he2020momentum}. However, for a better overall performance, we adopt batch size of $32$.

\subsection{Analysis}
\label{exp:analysis}
As illustrated in Fig.~\ref{fig:distribution}, we visualize the distribution of the composition features of left and right boundaries (the results of top and bottom boundaries are in similar distributions) from the CBLNet trained with and without C2C by reducing the features into two dimensions by t-SNE~\cite{van2008visualizing}. The location of the corresponding ground truth boundaries is denoted by color. The darker color means that the annotated boundary locates farther from the original input boundary. 
Without C2C, the features of the rare samples, whose ground truth is far away from the input image boundary, are mixed up with that of the normal samples. This means that the model without C2C can not properly depict the unseen rare samples, and seems to just project these samples randomly in the feature space. Hence, these samples follow the normal distribution of training set, which is the cause of quasi-trivial solutions.
With C2C, firstly, the rare samples with darker color clustered and the feature distribution aligns well with the location distribution of the regression targets, which shows the success of C2C. Secondly, the expected feature distribution leads to a better performance, which demonstrates the rationale of our observation and hypothesis that cropping boundaries locating nearly share similar composition patterns.

\section{Conclusions}

In this work, we study why existing cropping box regression algorithms tend to render quasi-trivial solution. We find that the crux lies in the limitation of representational capacity. Based on our observation, we make a hypothesis that cropping boundaries with similar annotations share similar composition patterns. Therefore, we propose to conduct composition feature clustering to enhance the composition representation. Firstly, by presenting CBLNet, the composition features of four cropping boundaries are obtained. Inspired by the idea of contrastive learning and deep clustering, we propose C2C to cluster the samples with similar annotations jointly with the normal optimization procedure. By training with C2C, our CBLNet can outperform other state-of-the-art regression-based image cropping models on the FCDB dataset even without extra data. Compared with other unbiased training baselines, the C2C also achieves superior performance with the little time cost. Our work reveals that cropping box regression networks actually can directly learn from the cropping box annotations, but how to train a network matters. Our work provides a new perspective on the task of cropping box regression and states that composition feature clustering can make a difference.

\section{Acknowledgements}
This work was funded by the DigiX Joint Innovation Center of Huawei-HUST.

\bibliography{aaai23}

\end{document}